%% file: arxiv_main.tex
\documentclass{article}

\usepackage{arxiv}
\usepackage{amsmath}
\usepackage[utf8]{inputenc} 
\usepackage[T1]{fontenc}    
\usepackage{hyperref}       
\usepackage{url}            
\usepackage{booktabs}       
\usepackage{amsfonts}       
\usepackage{nicefrac}       
\usepackage{microtype}      
\usepackage{lipsum}
\usepackage{graphicx}
\graphicspath{ {./images/} }
\usepackage{algorithm}
\usepackage{algorithmic}
\usepackage{graphicx}
\usepackage{subcaption}
\usepackage{soul}
\usepackage{tikz}
\usepackage{tabularx}
\usetikzlibrary{arrows.meta, positioning, shapes, fit}

\usepackage{hyperref}
\usepackage{wrapfig}
\hypersetup{
    colorlinks=true,
    linkcolor=blue,
    filecolor=blue,      
    urlcolor=blue,
}

\title{EvoTS: Evolutionary Transformer Search \\for Time Series Forecasting}

\author{
\begin{tabular}{cc}
AbdElRahman ElSaid & Damir Pulatov \\
\texttt{elsaida@uncw.edu} & \texttt{pulatovd@uncw.edu} \\
\multicolumn{2}{c}{University of North Carolina Wilmington} \\
\multicolumn{2}{c}{Wilmington, North Carolina, USA}
\end{tabular}
}

\begin{document}
\maketitle
\input{sections/00-abstract}


\input{sections/01-intro}
\input{sections/02-literature}
\input{sections/03-method}
\input{sections/04-experiments}
\input{sections/05-results}
\input{sections/06-discussion}

\bibliographystyle{ieeetr}
\bibliography{references}

\end{document}

%% file: sections/00-abstract.tex
\begin{abstract}

Evolutionary neural architecture design for multivariate time-series forecasting remains under-explored, with most approaches relying on fixed Transformer architectures despite substantial variation across tasks and forecasting settings. This paper introduces an evolutionary neural architecture search framework for discovering task-adaptive Transformer-like models for time-series forecasting (\textbf{\textsc{EvoTS}}). Architectures are encoded using a modular genome representation that enables flexible composition of attention, feed-forward, and projection components, while a repair mechanism enforces structural validity throughout the evolutionary process. This formulation allows effective exploration of a diverse architecture space without relying on hand-crafted design rules.

The proposed approach is evaluated on four benchmark datasets from the ETT family (ETTh1, ETTh2, ETTm1, and ETTm2) under multiple forecasting settings, including univariate-to-univariate, multivariate-to-univariate, and multivariate-to-multivariate prediction, with horizons of 96, 192, 336, and 720. In the multivariate-to-multivariate setting, the evolved architectures achieve competitive and, in several cases, improved mean squared error relative to a strong Transformer-based baseline. Additional analyses examine performance differences across forecasting settings and report wall-clock training time to provide a coarse indication of computational cost.

Overall, the results demonstrate that evolutionary search can effectively discover flexible and high-performing Transformer-like architectures for multivariate time-series forecasting within practical runtime constraints.

\end{abstract}

%% file: sections/01-intro.tex
\vspace{-.5cm}
\section{Introduction}
\label{sec:intro}

Accurate time-series forecasting is a critical component of many real-world systems, including energy management, industrial monitoring, and large-scale infrastructure control. 
Recent advances in Transformer-based architectures have significantly improved forecasting performance by enabling flexible modeling of long-range temporal dependencies and complex cross-variable interactions. 
As a result, Transformers and their variants have become a dominant paradigm in multivariate time-series forecasting.

Despite their success, most Transformer-based forecasting models rely on carefully hand-designed architectures. 
Choices such as tokenization strategy, attention formulation, block composition, and network depth are typically fixed a priori and tuned through extensive manual experimentation. 
While effective, this process is labor-intensive and often yields architectures that perform well on a narrow range of tasks or prediction horizons, limiting adaptability across datasets and forecasting regimes.

Evolutionary computation offers a principled alternative to manual architecture design. 
By treating neural architectures as evolvable structures and optimizing them through population-based search, evolutionary methods can explore diverse architectural compositions without committing to a single design template. 
This paradigm is particularly well-suited to neural architecture search problems where the design space is highly structured and the cost of evaluation is substantial.

In this work, we propose \textsc{EvoTS}\footnote{\url{https://github.com/a-elsaid/EVOTS.git}}, an evolutionary architecture search framework for Transformer-like time-series forecasting models.
The framework represents architectures as modular genomes composed of Transformer-inspired processing blocks and evolves their composition using a steady-state evolutionary algorithm with weight inheritance. 
Unlike approaches that introduce new attention mechanisms or training objectives, our method focuses on automating architectural composition under a fixed and widely adopted training protocol. 
All experiments follow the evaluation settings introduced in the iTransformer study~\cite{liu2024itransformer}, enabling direct and fair comparison to previous Transformer-based baselines.

We evaluate the proposed approach on four widely used ETT benchmark datasets (ETTh1, ETTh2, ETTm1, ETTm2) across multiple prediction horizons.
The evolved architectures consistently achieve strong performance in the multivariate-to-multivariate forecasting setting and match or exceed the strongest results reported in prior work under identical experimental conditions. 
Performance gains are particularly pronounced at longer prediction horizons, where fixed architectural designs are most challenged.

The main contributions of this paper are as follows:
    \textbf{\textit{i)}} We introduce a modular evolutionary architecture search framework for Transformer-like time-series forecasting models,
    \textbf{\textit{ii)}} we demonstrate that evolutionary search can automatically discover hybrid architectural compositions that outperform hand-designed Transformer variants on standard forecasting
    benchmarks, and
    \textbf{\textit{iii)}} we provide empirical analysis showing that the benefits of evolutionary architecture search are amplified in long-horizon and multivariate forecasting regimes.

%% file: sections/02-literature.tex
\section{Related Work}
\label{sec:literature}

\paragraph{Transformer-Based Time-Series Forecasting.}
Transformer architectures have been widely adopted for time-series forecasting due to their ability to model long-range temporal dependencies and complex interactions among variables~\cite{vaswani2017attention,zhou2021informer}. Early approaches primarily apply self-attention over temporal tokens, treating each time step as a token that aggregates all variables\cite{lim2021temporal, zhou2021informer}. 
Subsequent models introduce architectural modifications tailored to time-series structure.
Autoformer incorporates series decomposition and replaces self-attention with auto-correlation mechanisms to emphasize periodic patterns\cite{wu2021autoformer}.
Crossformer introduces segmented embeddings that jointly preserve temporal and cross-variable information~\cite{zhang2023crossformer}. 
More recently, iTransformer proposes inverting the attention mechanism by treating variables as tokens, enabling direct modeling of cross-variable relationships\cite{liu2024itransformer}.

Following the success of large pretrained foundation models which rely on Transformer based architectures, another research direction is to build a foundation model that tackles time series instead of natural language.
The advantage of such approaches is that pretrained models can work out of the box on any time series data and no training is required by end users.
Research into pretrained models for time series ranges from simply prompting~\cite{gruver2023large, xue2023promptcast} and fine-tuning~\cite{zhou2023one, jin2023time} existing general-purpose foundation models to training specialized models in which model architectures are modified to accommodate time series~\cite{rasul2023lag, goswami2024moment, das2024decoder, woo2024unified, ansari2024chronos, ansari2025chronos2}.
One of the drawbacks of such approaches is the limited amount of data available for training models of such a scale.
Pretrained models such as Chronos~\cite{ansari2024chronos} resort to data augmentation and the use of synthetic datasets in addition to real-world datasets while models such as ForecastPFN~\cite{dooley2023forecastpfn} and TabPFN-v2~\cite{tabpfnv2-time} resort to training models purely on synthetic data.

These works highlight the importance of architectural choices such as tokenization strategy, attention formulation, and block composition.
However, such choices are typically fixed a priori and manually tuned~\cite{elsken_neural_2019} for specific datasets and prediction horizons, limiting adaptability across forecasting regimes.

\paragraph{Neural Architecture Search and Neuroevolution.}
Neural Architecture Search (NAS) seeks to automate the design of neural network architectures and has been explored using a variety of optimization paradigms~\cite{elsken_neural_2019}. Evolutionary and neuroevolutionary approaches, including NEAT~\cite{stanley2002evolving}, EXAMM~\cite{ororbia2019investigating}, EXA-GP~\cite{murphy2024exa}, ANTS~\cite{elsaid2020ant}, CANTS~\cite{elsaid2021continuous,elsaid2025cg}, and NSGA-Net~\cite{lu2019nsga}, represent architectures as mutable structures and optimize them through population-based search. These methods are particularly well-suited to discrete and structured design spaces, supporting architectural variation, reuse, and incremental refinement through mechanisms such as mutation, recombination, and inheritance~\cite{stanley2002evolving,ororbia2019investigating,real2019regularized}.

\sloppy Alternative NAS paradigms include reinforcement-learning–based methods, such as controller-driven and weight-sharing approaches exemplified by ENAS~\cite{pham2018efficient}, as well as gradient-based techniques like DARTS~\cite{liu2018darts} that rely on continuous relaxations of discrete architectural choices. While these methods offer improved search efficiency, they typically operate over constrained architectural templates and are less naturally suited to search spaces involving heterogeneous modules and discrete structural variation.

\paragraph{NAS for Transformer Architectures.}
NAS has also been applied to Transformer architectures, primarily in natural language processing~\cite{so2019evolved} and computer vision. 
Prior work such as the Evolved Transformer and subsequent vision-oriented~\cite{chen2021searching} approaches restrict the search space to variations of a fixed Transformer template, focusing on parameters such as embedding dimension, number of attention heads, or layer depth. 
Other approaches expand the search space to include both self-attention and convolutional operations\cite{gong2022nasvit}, enabling hybrid designs while still enforcing predefined stage structures and tokenization schemes.

These studies demonstrate the feasibility of automating Transformer design, but generally emphasize hyperparameter optimization or selection among predefined modules, rather than evolving the internal composition of Transformer-like architectures under a fixed training and evaluation protocol~\cite{so2019evolved,gong2022nasvit}.

\sloppy\paragraph{Positioning of This Work.}
In contrast to prior NAS and Transformer-NAS approaches, this work focuses on evolutionary search over the internal composition of Transformer-like architectures for time-series forecasting. Architectures are represented as modular genomes~\cite{stanley2002evolving,ororbia2019investigating} composed of Transformer-inspired blocks, including multiple attention variants and convolutional token-mixing operations, and are evolved using a steady-state~\cite{de1999evolutionary} evolutionary algorithm with weight inheritance~\cite{real2019regularized,ororbia2019investigating}. 
Rather than introducing new attention mechanisms or task-specific heuristics, the proposed framework operates under a fixed and widely adopted training protocol~\cite{liu2024itransformer}, enabling direct comparison to existing Transformer-based forecasting models while automating architectural composition.

%% file: sections/03-method.tex
\section{Methodology}
\label{sec:method}
This section describes the \textsc{EvoTS} framework, including genome representation, evolutionary dynamics, and evaluation.

\subsection{Problem Setting}
We study time-series forecasting under three standard settings: univariate-to-univariate (S$\rightarrow$S), multivariate-to-univariate (M$\rightarrow$S), and multivariate-to-multivariate (M$\rightarrow$M) prediction. Given a historical input window of length $L$, the objective is to predict future values over a forecasting horizon $H \in \{96, 192, 336, 720\}$. Model performance is evaluated using mean squared error (MSE).

While all three settings are considered for completeness, we primarily emphasize the multivariate-to-multivariate setting, as it reflects the standard evaluation protocol adopted by recent Transformer-based time-series forecasting benchmarks.

\subsection{Evolutionary Architecture Search Framework}
We propose an evolutionary neural architecture search framework for discovering Transformer-like architectures tailored to time-series forecasting tasks. Rather than relying on a fixed, manually designed architecture, the framework explores a structured space of neural architectures using population-based evolutionary optimization, following established neuroevolutionary principles~\cite{stanley2019designing}.

The evolutionary process follows a steady-state paradigm~\cite{de1999evolutionary}, in which new candidate architectures are generated, evaluated, and immediately considered for inclusion in the population. Unlike generational evolution, steady-state evolution allows continuous refinement of the population and incremental improvement of candidate solutions, making it particularly well suited to computationally expensive evaluation settings such as neural architecture search.

\subsection{Genome Representation}
Each individual in the population is represented by a modular genome encoding a Transformer-like forecasting architecture. Modular and indirect genome representations have been shown to support scalable neuroevolution of deep neural networks~\cite{stanley2002evolving}. In our framework, the genome specifies (i) an input tokenization strategy, (ii) an ordered sequence of processing blocks, and (iii) an output projection head.

Architectural hyperparameters--including embedding dimensionality, number of attention heads, convolutional kernel sizes, and block ordering--are explicitly encoded in the genome, enabling fine-grained structural variation during evolution.

\paragraph{Evolution of Input and Output Heads}
Both the input tokenization module and the output projection head are encoded as part of the genome and are optimized through evolution. The input head determines how raw time-series observations are mapped into a shared token representation, while the output head defines how the learned token sequence is projected to the forecasting targets.

Evolution controls the \emph{type} of head selected as well as its internal architectural parameters, including encoder variants, kernel sizes, pooling strategies, and projection dimensionalities. However, heads are subject to structural constraints: exactly one tokenization module is applied at the input of the network, and exactly one output projection is applied at the end. Their position and multiplicity are fixed, and they are not interleaved with processing blocks.

Importantly, the internal composition of the heads is not evolved independently of the genome. Instead, head variants are selected deterministically based on compatibility with the tokenization strategy and the processing blocks specified by the genome. For example, frequency-domain encoders are instantiated only when required by frequency-based tokenization or processing blocks. This design enforces architectural consistency while allowing evolution to indirectly influence head structure through upstream design choices.

\paragraph{Tokenization}
Tokenization is an architectural choice applied once at model construction time. The tokenizer defines the token axis and initial representation on which all subsequent processing blocks operate. The tokenizer is selected from a fixed pool comprising time-wise, variable-wise, patch-based, and grouped cross-variable tokenization schemes.

All tokenization modules map raw time-series inputs to a shared tensor interface of shape $[B, T, D]$, where $B$ denotes batch size, $T$ the number of tokens, and $D$ the embedding dimension. This shared interface ensures compatibility across heterogeneous processing blocks.

Tokenization modules may optionally incorporate skip connections from the raw input to preserve low-level information during encoding. These skip connections are local to the tokenization stage and are not dynamically inserted between processing blocks.

\paragraph{Architectural Constraints}
To bound the search space while preserving architectural flexibility, genomes are constrained to form architectures consisting of a single tokenization stage followed by a sequence of processing blocks and a final output projection. All processing blocks operate on the shared token representation produced by the tokenizer.

Dimensional compatibility between blocks is enforced by construction, and projection layers are restricted to appear only at the input and output boundaries of the architecture. Intermediate retokenization, token-space transitions, or dynamic changes in token structure are reserved for future work.

\paragraph{Supported Block Types}
The modular genome supports a heterogeneous set of Transformer-inspired and token-mixing processing blocks. Attention-based blocks include standard multi-head self-attention as well as inverted and cross-variable attention variants commonly used in time-series forecasting. In addition to attention mechanisms, the search space includes convolutional token-mixing blocks that operate along the token dimension, providing an alternative means of capturing local temporal dependencies.

Multiple block types may coexist within a single genome. Compatibility between heterogeneous blocks is achieved by enforcing a common token interface rather than by introducing intermediate tokenization or projection layers. Residual (skip) connections are applied within blocks as part of their standard definitions to support optimization stability and effective gradient flow.

Although frequency-domain processing blocks and encoders are available within the architectural vocabulary, they are instantiated only when explicitly required by the genome-defined tokenization or block types. As a result, frequency-domain encoders are not universally present and may be entirely absent from evolved architectures if evolution does not favor such representations.

Output heads support both single-target and multi-target forecasting through flexible projection and pooling mechanisms. While the current block vocabulary reflects components commonly used in Transformer-based forecasting models, the genome design is block-agnostic and readily extensible to additional block types.

\begin{figure}[t]
    \centering
    \includegraphics[width=0.95\linewidth]{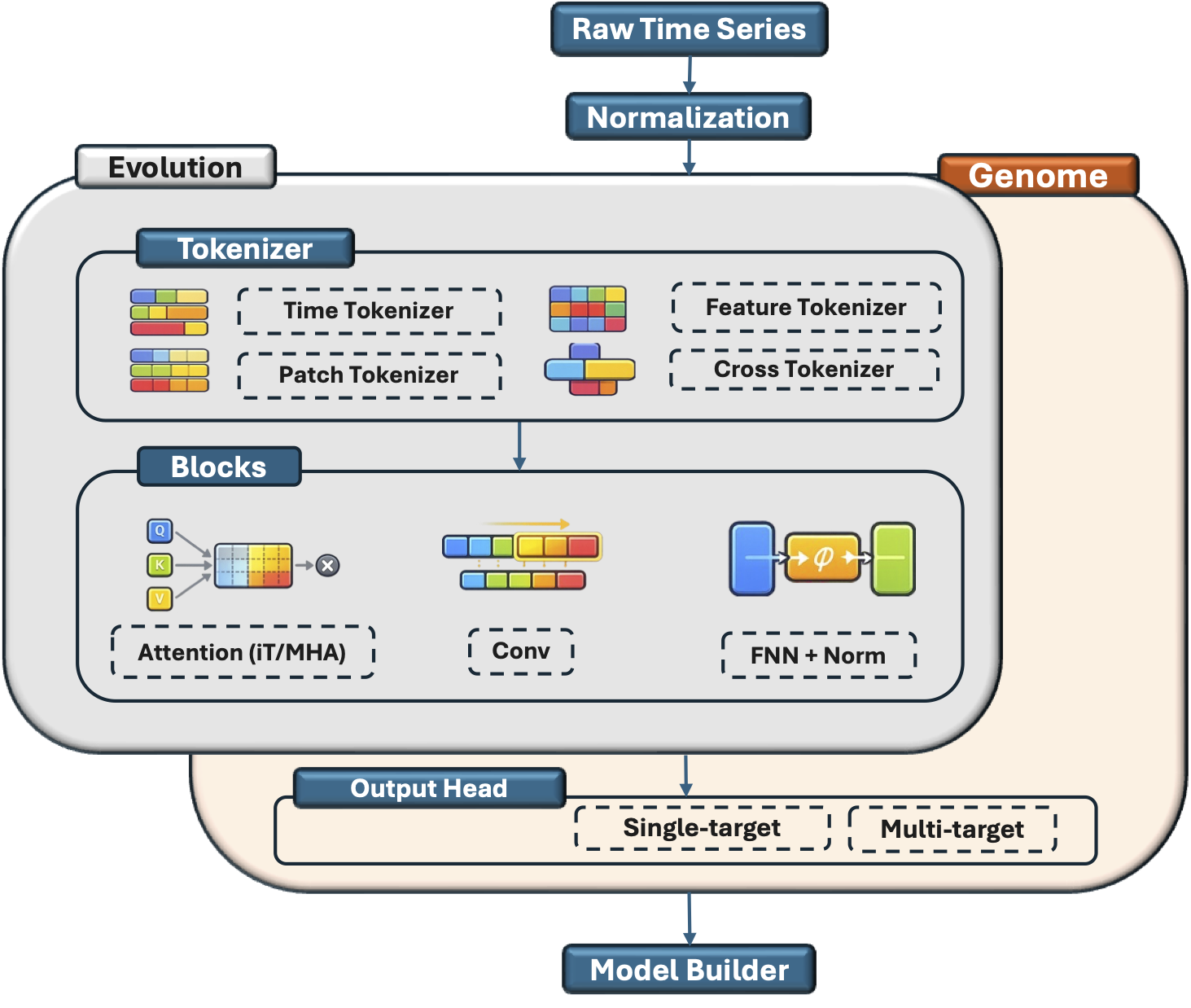}
    \caption{\centering \textit{High-level schematic of the modular architecture search space.}\vspace{-0.75cm}}
    \label{fig:architecture}
\end{figure}

Figure~\ref{fig:architecture} provides a high-level overview of the modular architecture search space.

\subsection{Population Initialization}
The initial population is constructed using a combination of randomly sampled genomes and a small number of seed architectures. Seed genomes correspond to simple, valid Transformer-like configurations and serve to stabilize the early stages of evolutionary search. Randomly initialized individuals are included to promote architectural diversity and reduce bias toward any single design pattern.

\subsection{Structural Repair Mechanism}
Evolutionary variation operators may produce invalid or ill-formed architectures. To ensure executability, a deterministic repair mechanism is applied after mutation or recombination. The repair procedure enforces dimensional compatibility between modules, ensures the presence of required components such as tokenization and output projections, and maintains valid computational connectivity.

Repair does not introduce additional stochasticity and does not alter the evolutionary objective, enabling unrestricted exploration of the search space while guaranteeing that all evaluated individuals correspond to valid neural networks.

\subsection{Variation Operators}
New architectures are generated using evolutionary variation operators adapted to the modular genome representation. Mutation operators include block insertion, removal, replacement, and parameter perturbation. Recombination operators combine architectural components from two parent genomes.

When architectural compatibility permits, offspring architectures may optionally reuse learned parameters from parent models. Such parameter reuse can improve sample efficiency in evolutionary architecture search, particularly when training deep neural networks from scratch. 
In the experiments reported in this work, weight inheritance was not active during search; this capability is included in the framework for future use.

\subsection{Evolutionary Search Dynamics}
The evolutionary search begins by initializing a population of candidate architectures sampled from the defined genome space. At each evolutionary step, parent individuals are selected using tournament selection. Offspring architectures are generated through mutation and recombination, followed by deterministic repair to ensure structural validity.

Each offspring is trained and evaluated to obtain its fitness value. In the steady-state setting, evaluated offspring are immediately compared against existing population members and replace inferior individuals if they achieve superior fitness. This process continues until a predefined termination criterion is met, such as a maximum number of architecture evaluations.

Algorithm~\ref{alg:evolution} provides a concise summary of the steady-state evolutionary architecture search procedure used in this work.

\subsection{Fitness Evaluation}
Each candidate architecture is trained using a fixed training protocol and evaluated on a validation set to compute its fitness. Fitness is defined as the validation mean squared error for the corresponding forecasting task and horizon. Training is performed for a fixed number of epochs with early stopping based on validation performance.

Evaluations of individuals are independent and can be executed in parallel, enabling efficient utilization of available computational resources. The evolutionary algorithm itself remains agnostic to the underlying hardware configuration.

\subsection{Selection and Replacement}
Selection is performed using tournament selection, favoring individuals with lower validation error while maintaining population diversity. In the steady-state setting, newly evaluated offspring replace the worst-performing individuals in the population if they achieve superior fitness.

\subsection{Final Model Selection}
The architecture with the lowest validation MSE observed during evolutionary search is selected, and all reported results correspond to this model.

\begin{algorithm}[t]
\caption{Steady-State Evolutionary Architecture Search}
\label{alg:evolution}
\begin{algorithmic}[1]
\REQUIRE Population size $N$, maximum evaluations $E_{\max}$
\STATE Initialize population $\mathcal{P} = \{g_1, \dots, g_N\}$ with random valid genomes
\FORALL{$g \in \mathcal{P}$}
    \STATE Train model defined by $g$ and evaluate fitness $f(g)$
\ENDFOR
\STATE $e \leftarrow N$
\WHILE{$e < E_{\max}$}
    \STATE Select parent(s) $g_p$ from $\mathcal{P}$ via tournament selection
    \STATE Generate offspring genome $g_o$ via mutation and/or recombination
    \STATE Apply repair operator to $g_o$ to ensure structural validity
    \IF{architectural compatibility permits}
        \STATE Initialize $g_o$ with inherited parameters from $g_p$
    \ENDIF
    \STATE Train model defined by $g_o$ and evaluate fitness $f(g_o)$
    \STATE $e \leftarrow e + 1$
    \STATE Identify worst individual $g_w \in \mathcal{P}$
    \IF{$f(g_o) < f(g_w)$}
        \STATE Replace $g_w$ with $g_o$ in $\mathcal{P}$
    \ENDIF
\ENDWHILE
\STATE \textbf{return} best genome $g^* \in \mathcal{P}$
\end{algorithmic}
\end{algorithm}

%% file: sections/04-experiments.tex
\section{Experimental Setup}
\label{sec:exp}

Unless otherwise stated, all datasets, input lengths, prediction horizons, training procedures, and evaluation protocols follow the experimental
configuration described in the iTransformer study~\cite{liu2024itransformer}.
This ensures direct comparability between our results and previously reported Transformer-based baselines.

\subsection{Datasets}
We evaluate the proposed method on four widely used benchmarks from the Electricity Transformer Temperature (ETT) dataset family: ETTh1, ETTh2, ETTm1, and ETTm2. 
These datasets consist of multivariate time-series measurements collected from electricity transformers, including load, oil temperature, and related operational variables. 
Following standard practice, ETTh datasets are sampled at an hourly resolution, while ETTm datasets are sampled at a 15-minute resolution.

All datasets are split chronologically into training, validation, and test sets using the standard partitioning protocol adopted by prior Transformer-based forecasting studies.  No future information is used during training or validation.

\subsection{Forecasting Settings}
Experiments are conducted under three forecasting settings:
\begin{itemize}
    \item \textbf{Univariate-to-univariate (S$\rightarrow$S)}: a single input
    variable is used to predict the same variable in the future.
    \item \textbf{Multivariate-to-univariate (M$\rightarrow$S)}: all input
    variables are used to predict a single target variable.
    \item \textbf{Multivariate-to-multivariate (M$\rightarrow$M)}: all input
    variables are jointly predicted.
\end{itemize}

For all settings, forecasting horizons of $H \in \{96, 192, 336, 720\}$ are considered. 
The multivariate-to-multivariate setting is emphasized, as it aligns with the standard evaluation protocol used by recent Transformer-based forecasting models.

\subsection{Data Preprocessing}
Input time series are normalized using per-variable z-score normalization, where statistics are computed from the training set only and applied consistently to validation and test data. Sliding windows with fixed input length ($L=384$) are used to construct training samples, following established practice in Transformer-based forecasting.

\subsection{Baselines}
We compare the proposed evolutionary approach against iTransformer, a strong Transformer-based baseline for multivariate time-series forecasting.
iTransformer results are taken directly from the original publication and reported under identical datasets, prediction horizons, and evaluation metrics. 
Although the iTransformer paper reports several competitive architectures, iTransformer is used here as a representative Transformer baseline for direct comparison.

\subsection{Training and Evolution Protocol}
All candidate architectures are trained using a uniform optimization procedure to ensure fair comparison. 
Models are optimized with AdamW using a fixed learning rate of $1\times10^{-4}$, weight decay $3\times10^{-4}$, and batch size 32. 
Early stopping is enabled based on validation MSE with a patience of 10. Each evolutionary evaluation consists of a short initial training phase (20 epochs). Fitness is defined as validation MSE.

The evolutionary search follows a steady-state strategy with a population size of 20 and a maximum budget of 300 architecture evaluations. 
Tournament selection (size 2) is used, with mutation and crossover rates of 0.3 and 0.7, respectively. 
All experiments use a fixed random seed to ensure reproducibility. 
The initial population includes a small number of shallow seed architectures corresponding to standard Transformer variants (e.g., iTransformer- and PatchTST-style configurations with 4 layers), while the remaining individuals are randomly initialized to preserve architectural diversity.

\subsection{Runtime Measurement and Hardware}
To provide a coarse indication of computational cost, we report wall-clock training time for individual model evaluations. Each candidate architecture is trained and validated on a single GPU. 
Evolutionary evaluations are executed in parallel using three independent processes, each spawning and training one model at a time.

All experiments were conducted on a Linux-based system equipped with three NVIDIA RTX A6000 GPUs (48~GB memory each) and an AMD Ryzen Threadripper PRO 5995WX CPU.
Runtime measurements reported in Section~\ref{sec:result} reflect wall-clock performance under this hardware configuration.

%% file: sections/05-results.tex
\section{Results}
\label{sec:result}

\subsection{Multivariate-to-Multivariate Forecasting Performance}

\begin{figure*}[t]
    \centering

    \begin{subfigure}{0.48\textwidth}
        \centering
        \includegraphics[width=\linewidth]{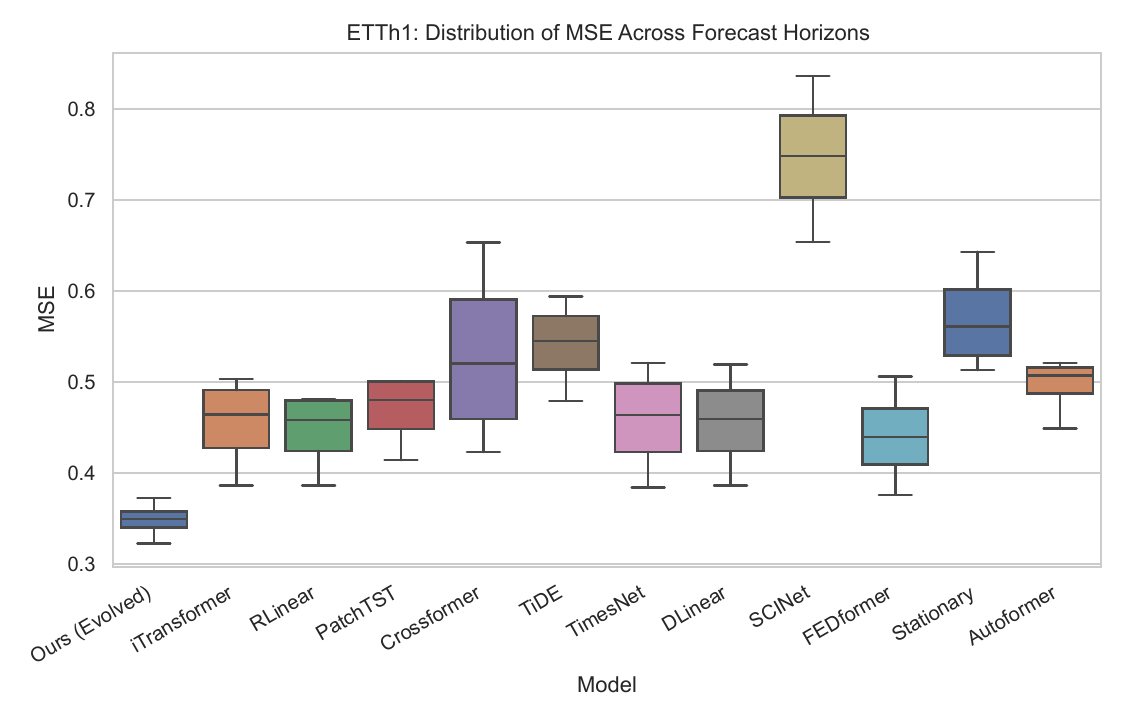}
        \caption{ETTh1}
    \end{subfigure}
    \hfill
    \begin{subfigure}{0.48\textwidth}
        \centering
        \includegraphics[width=\linewidth]{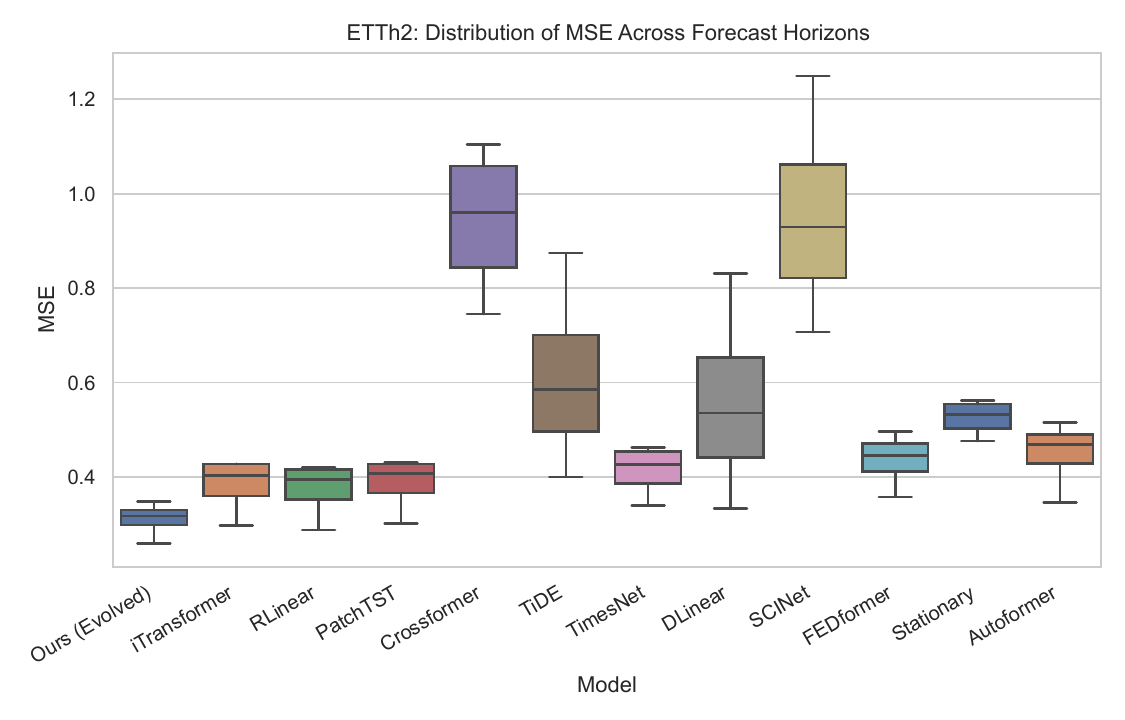}
        \caption{ETTh2}
    \end{subfigure}

    \vspace{0.5em}

    \begin{subfigure}{0.48\textwidth}
        \centering
        \includegraphics[width=\linewidth]{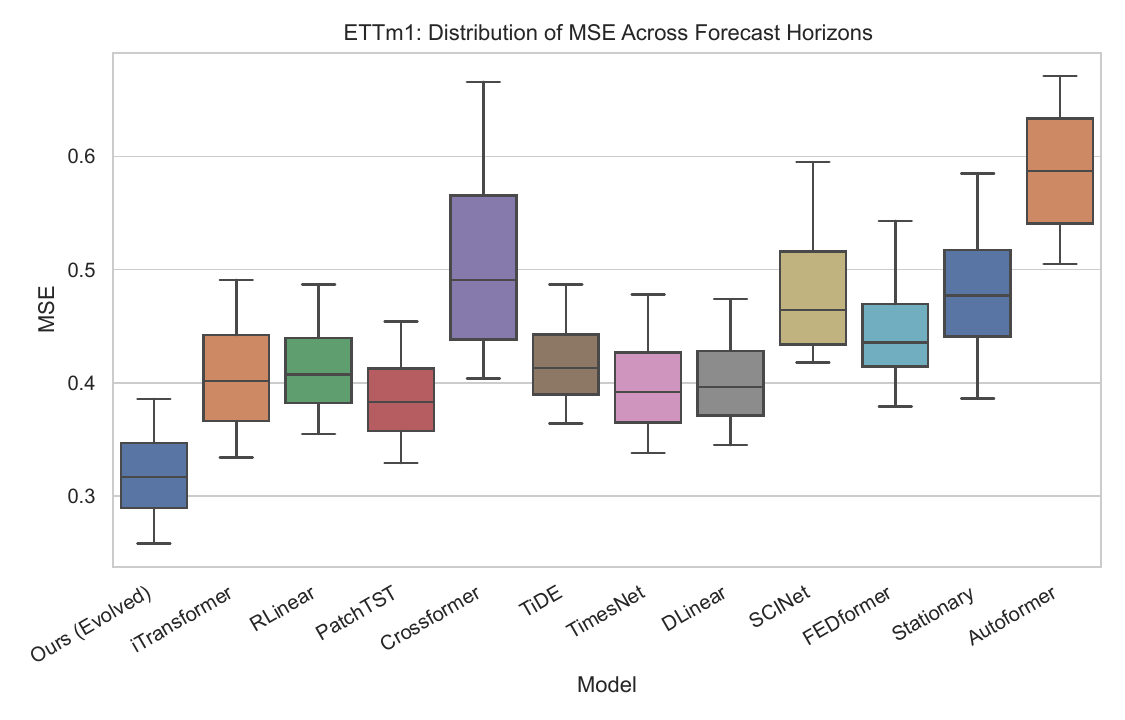}
        \caption{ETTm1}
    \end{subfigure}
    \hfill
    \begin{subfigure}{0.48\textwidth}
        \centering
        \includegraphics[width=\linewidth]{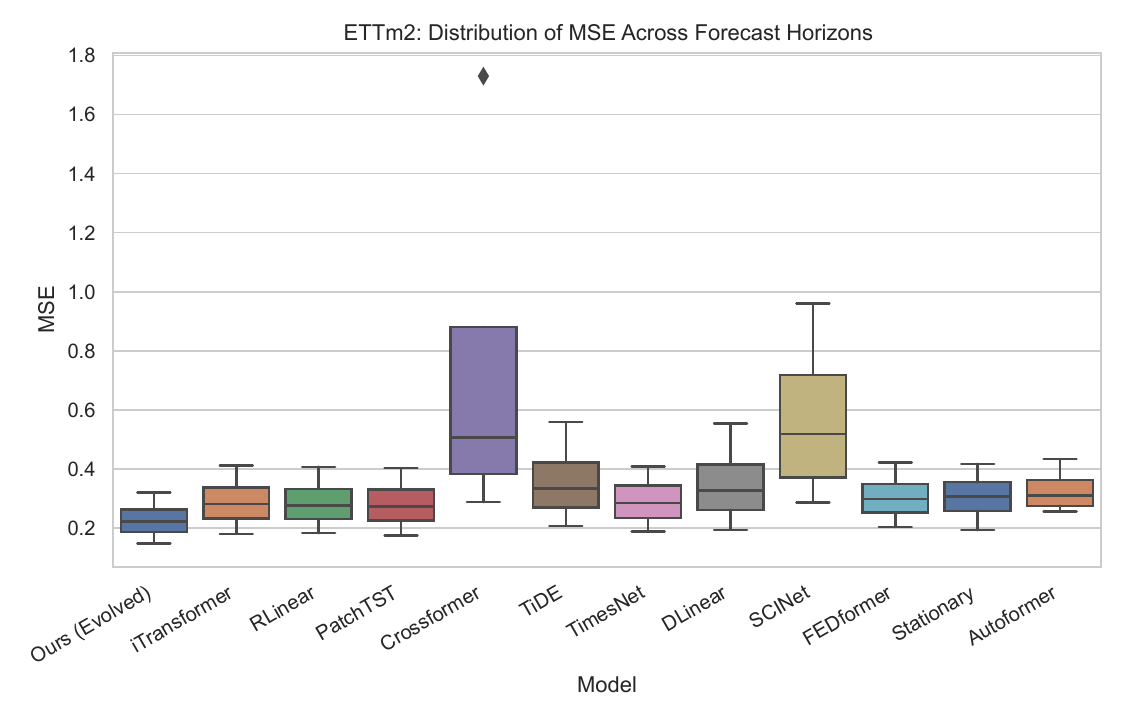}
        \caption{ETTm2}
    \end{subfigure}

    \caption{\centering
    Distribution of MSE across forecasting horizons
    ($96$, $192$, $336$, $720$) for $M\rightarrow M$ forecasting.
    Each box summarizes performance variability across horizons for a given
    model.
    }
    \label{fig:mmm_boxplots}
\end{figure*}

Table~\ref{tab:mmm_results} reports mean squared error (MSE) for the multivariate-to-multivariate (M$\rightarrow$M) forecasting setting on four ETT
benchmark datasets (ETTh1, ETTh2, ETTm1, ETTm2) with prediction horizons of 96, 192, 336, and 720. 
Results for iTransformer are taken directly from the original publication and correspond to identical datasets, horizons, and evaluation protocols. As is standard in NAS, validation MSE serves as both the fitness criterion during search and the reported performance metric.   
Figure~\ref{fig:mmm_boxplots} visualizes the distribution of forecasting errors across horizons and datasets for the same experimental setting.

Across all evaluated datasets and prediction horizons, the architectures discovered by the proposed evolutionary framework achieve MSE that is lower
than or comparable to the strongest results reported for the ETT benchmarks in the iTransformer study~\cite{liu2024itransformer}, under identical datasets, horizons, and evaluation protocols. Performance gains are modest at shorter horizons and become more pronounced as the prediction horizon increases. 
For example, on ETTh1, MSE is reduced from 0.376 to 0.3224 at horizon 96 and from 0.481 to 0.3724 at horizon 720, corresponding to relative error reductions of 14.2\% and 22.6\%, respectively. Similar trends are observed on ETTh2, ETTm1, and ETTm2, indicating that the benefits of evolutionary architecture search are particularly evident in longer-horizon forecasting scenarios.

The iTransformer study~\cite{liu2024itransformer} evaluates a range of Transformer-based forecasting models, including Autoformer, FEDformer,
Informer, and Crossformer, and reports that different architectures achieve the best performance depending on dataset and prediction horizon. 
In particular, iTransformer does not uniformly dominate all baselines across the ETT benchmarks but serves as a strong and representative modern Transformer baseline under a consistent evaluation protocol.

In this context, our comparison against iTransformer should be interpreted as a controlled reference rather than a comparison to the strongest possible model in all settings. 
The proposed evolutionary approach consistently improves upon iTransformer across datasets and horizons, and the distributional comparisons shown in Figure~\ref{fig:mmm_boxplots} indicate that these gains are robust across short- and long-range forecasting regimes. 
While direct numerical comparison with all baselines reported in~\cite{liu2024itransformer} is beyond the scope of this study, the observed improvements relative to iTransformer suggest that the evolved architectures are competitive with, and in many cases comparable to, the best-performing Transformer variants reported in prior work.

\begin{table}[t]
\centering
\caption{\centering MSE comparison under multivariate-to-multivariate
(M$\rightarrow$M) forecasting. Lower is better.}
\label{tab:mmm_results}
\begin{tabular}{lcccc}
\toprule
\textbf{Model} & \textbf{96} & \textbf{192} & \textbf{336} & \textbf{720} \\
\midrule
\multicolumn{5}{c}{\ul{\textbf{ETTh1}}} \\
Ours (Evolved) & \textbf{0.3224} & \textbf{0.3462} & \textbf{0.3527} & \textbf{0.3724} \\
iTransformer & 0.3760 & 0.4200 & 0.4200 & 0.4810 \\
\midrule
\multicolumn{5}{c}{\ul{\textbf{ETTh2}}} \\
Ours (Evolved) & \textbf{0.2591} & \textbf{0.3109} & \textbf{0.3245} & \textbf{0.3485} \\
iTransformer & 0.2880 & 0.3740 & 0.4150 & 0.4200 \\
\midrule
\multicolumn{5}{c}{\ul{\textbf{ETTm1}}} \\
Ours (Evolved) & \textbf{0.2580} & \textbf{0.2999} & \textbf{0.3336} & \textbf{0.3858} \\
iTransformer & 0.3290 & 0.3670 & 0.3990 & 0.4540 \\
\midrule
\multicolumn{5}{c}{\ul{\textbf{ETTm2}}} \\
Ours (Evolved) & \textbf{0.1470} & \textbf{0.2000} & \textbf{0.2442} & \textbf{0.3196} \\
iTransformer & 0.1750 & 0.2410 & 0.3050 & 0.4020 \\
\bottomrule
\end{tabular}
\end{table}

\subsection{Univariate and Multivariate-to-Univariate Performance}

For completeness, Table~\ref{tab:other_results} reports results for the
univariate-to-univariate (S$\rightarrow$S) and multivariate-to-univariate
(M$\rightarrow$S) forecasting settings. In the univariate setting, the evolved
architectures achieve consistently strong performance across all datasets and
horizons, confirming that the search space remains effective even in simple
forecasting regimes.

In contrast, performance in the multivariate-to-univariate setting exhibits
higher variance and, in several cases, degradation relative to the
multivariate-to-multivariate setting for the same evolved architectures. This behavior suggests that the current
architectural design and output-head configurations are better suited to joint
multivariate prediction, where shared representations across variables can be
exploited more effectively. These results are therefore presented as diagnostic
evidence rather than primary benchmarks.

\begin{table}[t]
\centering
\caption{\centering
MSE results for univariate-to-univariate (S$\rightarrow$S) and
multivariate-to-univariate (M$\rightarrow$S) forecasting.}
\label{tab:other_results}
\begin{tabular}{llcccc}
\toprule
\textbf{Dataset} & \textbf{Setting} & \textbf{96} & \textbf{192} & \textbf{336} & \textbf{720} \\
\midrule
ETTh1 & S$\rightarrow$S & 0.0604 & 0.0725 & 0.0639 & 0.0713 \\
ETTh1 & M$\rightarrow$S & 0.1303 & 0.1690 & 0.1722 & 0.1538 \\
\midrule
ETTh2 & S$\rightarrow$S & 0.1063 & 0.1459 & 0.1575 & 0.1598 \\
ETTh2 & M$\rightarrow$S & 0.4313 & 0.5676 & 0.4398 & 0.3702 \\
\midrule
ETTm1 & S$\rightarrow$S & 0.0309 & 0.0493 & 0.0678 & 0.0972 \\
ETTm1 & M$\rightarrow$S & 0.3665 & 0.2725 & 0.3830 & 0.4227 \\
\midrule
ETTm2 & S$\rightarrow$S & 0.0671 & 0.1002 & 0.1310 & 0.1919 \\
ETTm2 & M$\rightarrow$S & 0.7103 & 0.5734 & 0.7715 & 0.6129 \\
\bottomrule
\end{tabular}
\end{table}

\subsection{Computational Cost and Runtime Analysis}

To assess computational feasibility, we measure wall-clock training time and CPU
utilization for evolutionary evaluations. We report approximate average CPU parallelism, estimated as
\[
\text{avg\_cores} \approx \frac{\text{user} + \text{sys}}{\text{real}}.
\]

Aggregated runtime statistics (Table~\ref{tab:runtime_summary}) show that average CPU utilization remains stable at approximately 3.6--3.9 cores across most configurations, largely independent of prediction horizon or dataset.
This indicates that the computational cost of the evolutionary procedure scales predictably and does not introduce excessive overhead beyond model training.

Longer prediction horizons are occasionally associated with reduced wall-clock time, which may reflect smoother long-term dynamics and faster convergence in some settings.
However, several multi-output configurations at horizon 720 exhibit increased runtime, reflecting the higher computational cost of producing large output tensors and reduced gradient signal-to-noise during optimization.
Overall, these results demonstrate that the proposed evolutionary search remains computationally tractable for realistic time-series forecasting workloads.

\begin{table}[t]
\centering
\small
\caption{\centering
Runtime statistics summary across all datasets (hours).}
\label{tab:runtime_summary}
\begin{tabular}{lcccc}
\toprule
\textbf{Setting} & \textbf{Median Real} & \textbf{Min Real} & \textbf{Max Real} & \textbf{Avg Cores} \\
\midrule
S$\rightarrow$S & 6.6 & 2.2 & 11.0 & 3.7 \\
M$\rightarrow$S & 10.5 & 2.0 & 14.0 & 3.7 \\
M$\rightarrow$M & 11.4 & 2.2 & 54.9$^\dagger$ & 3.8 \\
\bottomrule
\textbf{Horizon}  & 96    & 192   & 336   & 720 \\
\bottomrule
\end{tabular}
\end{table}

\subsection{Architecture Analysis of Evolved Models}
\label{subsec:arch_analysis}

\begin{table}[t]
\centering
\caption{\centering
Summary of architectural characteristics of best evolved models across datasets and prediction horizons ($M \rightarrow M$).}
\label{tab:arch_summary}
\footnotesize
\begin{tabular}{lcccc}
\toprule
\textbf{Characteristic} & \textbf{96} & \textbf{192} & \textbf{336} & \textbf{720} \\
\midrule
Avg. number of blocks        
& 6.0 (4--8) 
& 6.3 (4--8) 
& 5.0 (4--6) 
& 5.5 (4--7) \\

Inverted attention present   
& 50\% 
& 75\% 
& 75\% 
& 100\% \\

Standard attention present   
& 50\% 
& 75\% 
& 50\% 
& 75\% \\

Convolution blocks present   
& 75\% 
& 75\% 
& 75\% 
& 50\% \\

FFT encoder in input head    
& 25\% 
& 25\% 
& 0\% 
& 0\% \\

Cross-variable head enabled  
& 0\% 
& 0\% 
& 0\% 
& 0\% \\
\bottomrule
\end{tabular}
\end{table} 

To gain insight into the structures discovered by the evolutionary search, we analyze the best fine-tuned genomes produced across datasets and prediction horizons. 
Rather than presenting full genome specifications, we summarize key architectural characteristics of the evolved models, including tokenizer choice, block composition, and the configuration of input and output heads.
Representative architectural patterns are summarized in Table~\ref{tab:arch_summary}, where values report mean and range across four datasets per horizon.

Across all datasets and horizons, the evolved architectures consistently combine multiple Transformer-inspired processing mechanisms within a shared token space.
Inverted-attention, standard self-attention, and convolutional token-mixing blocks frequently co-exist within the same model, indicating that hybrid designs are systematically favored over architectures relying on a single inductive bias. 
This suggests that the evolutionary process benefits from balancing global dependency modeling (attention-based blocks) with local pattern extraction (convolutional blocks).

Tokenizer selection generally aligns with established architectural families. 
The majority of best-performing models adopt variable-wise tokenization, consistent with iTransformer-style designs, while patch-based or cross-variable tokenization does not appear in the top-performing single-stage architectures for the multivariate-to-multivariate setting considered here. 
Nevertheless, within a fixed tokenizer, substantial diversity is observed in block ordering and composition, highlighting the importance of internal architectural arrangement beyond high-level family selection.

All architectures analyzed in this study operate under a single-stage constraint, with a fixed tokenization applied once at the input. 
The present experiments operate under a single-stage constraint. Multistage architectures with intermediate retokenization are not included in the current search space and are left for future work.

Input and output heads are configured to match the selected tokenization scheme and are not evolved independently of the block structure. 
In particular, variable-wise tokenization is consistently paired with a variable head, while cross-variable heads are not activated in the reported models. 
FFT-based encoder variants appear only sporadically within input heads at shorter horizons and are absent at longer horizons, suggesting that frequency-domain preprocessing is not universally beneficial under the fixed training protocol used in this study.

Overall, the architectural analysis indicates that evolutionary search favors compact, single-stage Transformer-like models that flexibly combine attention and convolutional token-mixing mechanisms under a shared token interface. 
These results highlight the role of internal block composition and ordering--rather than architectural depth or token-space transitions--as key drivers of performance in the explored forecasting regimes.

%% file: sections/06-discussion.tex
\section{Discussion}
\label{sec:discussion}

The results presented in this work demonstrate that evolutionary architecture search provides a practical and effective alternative to
manual Transformer design for time-series forecasting. 
Across four ETT benchmarks and multiple prediction horizons, the proposed framework consistently discovers architectures that achieve strong performance
relative to established Transformer-based references evaluated under identical experimental protocols.

\sloppy A key empirical observation is that performance gains increase with the prediction horizon. At short horizons, most modern Transformer\-based models already perform well, as forecasting is dominated by local temporal continuity and short-range dependencies. 
As the prediction horizon grows, long-range temporal structure and cross-variable interactions become increasingly important, exposing limitations of
fixed, hand-designed architectures. 
In this regime, the flexibility of evolutionary search becomes more advantageous, enabling the discovery of architectural compositions that better balance global and local modeling requirements.

Architectural analysis reveals that the best-performing evolved models do not converge to a single canonical Transformer design. 
Instead, they frequently integrate multiple Transformer-inspired processing mechanisms within a shared token representation. 
Inverted attention, standard self-attention, and convolutional token-mixing blocks often co-exist within the same architecture. 
This hybrid composition suggests that no single inductive bias is sufficient across datasets and horizons, and that combining complementary mechanisms enables more robust forecasting behavior. 
The evolutionary process naturally favors such compositions by retaining only those components that contribute to fitness improvements.

Another consistent finding is the dominance of variable-wise tokenization in the multivariate-to-multivariate forecasting setting.
All best-performing architectures analyzed in this study operate on variable tokens and employ a single-stage processing pipeline. 
The present experiments operate under a single-stage constraint. Multistage architectures with intermediate retokenization are not included in the current search space and are left for future work.
The absence of patch-based or cross-variable tokenization in the top-performing models should therefore be interpreted in the context of the imposed single-stage design rather than as a general limitation of such tokenization strategies.

The strongest performance gains are observed in the multivariate-to-multivariate forecasting regime. 
Joint prediction of all variables provides a richer learning signal and allows architectures to exploit shared representations across correlated time series. 
In contrast, univariate and multivariate-to-univariate settings impose stronger output constraints, reducing the potential benefits of architectural adaptation. These findings indicate that evolutionary architecture search is particularly well-suited to forecasting problems with high structural complexity and strong cross-variable dependencies.

From a computational perspective, the steady-state evolutionary design enables efficient use of parallel resources while avoiding disruptive
population turnover. 
Runtime measurements indicate stable CPU utilization across datasets and horizons, with increased cost primarily associated with larger output dimensionality at long horizons. 
Despite the computational expense of evolutionary evaluation, the results demonstrate that the approach remains tractable for realistic forecasting workloads and benefits from weight inheritance and asynchronous evaluation.

Overall, the results suggest that the primary advantage of evolutionary search in this context lies not in increasing model size or depth, but
in automating the discovery of effective architectural compositions.
By removing the need to commit to a fixed design template, the framework adapts Transformer-like architectures to the statistical structure of the forecasting task and prediction horizon.

\subsection*{Future Work}

Several directions for future research naturally follow from this study.
First, extending the evolutionary search to multistage architectures with intermediate retokenization would allow exploration of transitions
between variable-wise, patch-based, and cross-variable token spaces within a single model, which may benefit data with heterogeneous temporal scales.

Second, expanding the block vocabulary to include additional Transformer-derived and non-Transformer modules could further enrich the
search space and enable discovery of novel hybrid architectures. 
This includes more expressive convolutional operators, frequency-domain processing blocks (e.g., FFT-based transformations), and adaptive attention mechanisms, enabling joint exploration of time- and frequency-domain representations.

Third, incorporating multi-objective fitness criteria that jointly optimize forecasting accuracy, computational cost, and model complexity would allow the evolutionary process to balance performance and efficiency, which is particularly relevant for resource-constrained deployment. 
Finally, evaluating the framework on a broader range of real-world datasets would provide deeper insight into its generality and robustness.